\documentclass[cameraready]{acmsiggraph}                     

\usepackage{times}[11]
\usepackage{amssymb}

\usepackage{graphicx}

\usepackage{parskip}

\usepackage[labelfont=bf,textfont=it]{caption}

\usepackage{bibspacing}
\onlineid{0}

\title{The Attentive Perceptron}

\author{Raphael A. Pelossof\\
Columbia University\\
{\tt\small pelossof@cs.columbia.edu}
\and
Zhiliang Ying\\
Columbia University\\
{\tt\small zying@stat.columbia.edu}
}


\def\SIN_FIG1{
\includegraphics[width=6.4in]{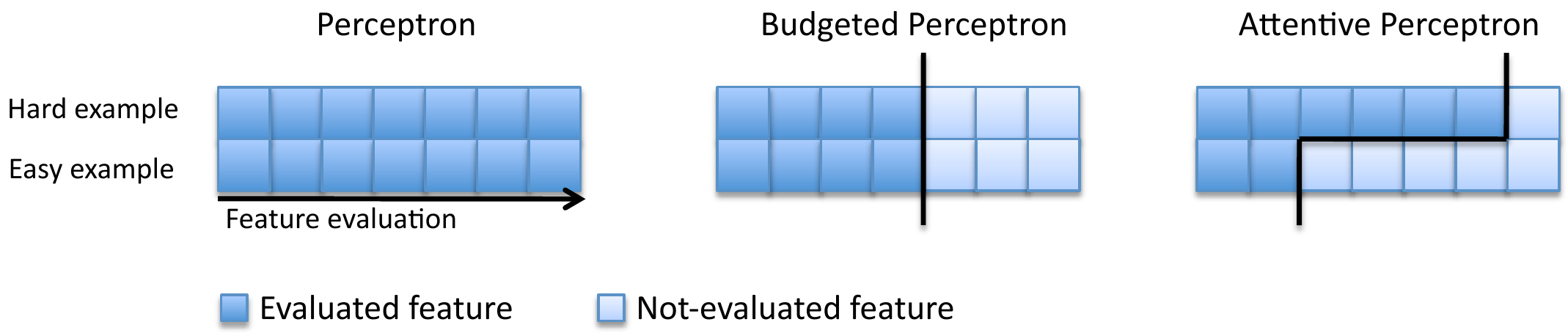}
\caption{\textbf{The Attentive Perceptron} 
adaptively allocates computational effort according to how hard an example is to classify. While the traditional \textit{Perceptron} evaluates all the features for all the examples, a \textit{Budgeted Perceptron} can only evaluate a constant number of features which is limited by the hard budget. From a budgeted learning point of view, the \textit{Attentive Perceptron} adaptively allocates computation while maintaining an average budget. Therefore easily classifiable examples are filtered after having evaluated a few of their features, whereas hard to classify examples have the majority of their features evaluated.}
\label{fig:attentive-perceptron}
}

\def\MNIST_TBL{
\begin{table*}[ht]
\caption{MNIST Dataset. Each digit classifier contains $1000$ features. Curtailed Online Boosting maintains similar accuracy as a fully updated $1000$ feature classifier, although it computes on average fewer than $200$ features.
}
\vspace{0.1in} 
\label{tbl:mnist_err}
\begin{tabular*}{0.95\textwidth}{@{\extracolsep{\fill}}|c|c|c|c|c|c|c|c|c|c|c|}
\hline
\multicolumn{11}{|c|}{Classification Error in \%} \\
\hline
Digit & 0 & 1 & 2 & 3 & 4 & 5 & 6 & 7 & 8 & 9 \\
\hline
AdaBoost, 1000 features 			 & 0.31 & 0.19 & 0.8 &0.89& 0.9 &1 &0.47& 0.79& 1.6 &1.3 \\
\hline\hline
Online Boosting, 1000 features 		 & 0.35 & 0.27 & 0.79 &1& 0.85& 1& 0.55& 0.97& 1.8& 1.4 \\
Curtailed Online Boosting & 0.43 & 0.31 & 0.84 &1.1 &1.1& 0.99& 0.74& 0.94& 1.4& 1.4 \\   
Online Boosting, 200 features &  0.47 
& 0.28 
& 1.55 
& 1.69 
& 1.60 
& 1.61 
& 0.85 
& 1.31 
& 3.61 
& 2.34 \\ 
\hline
\multicolumn{11}{|c|}{Curtailed Online Boosting Computational Efficiency} \\
\hline
Avg. Num. Features Evaluated & 145 &  175 &  135 &  153  & 108 &  141 &  107  & 101 &  202  & 150 \\
Speedup & 7  &   6  &   7 &    7 &    9  &   7  &   9  &  10 &    5  &   7 \\
\hline
\end{tabular*}
\end{table*}
}

\begin{document}

\teaser{\SIN_FIG1
}


\maketitle
\begin{abstract}
We propose a focus of attention mechanism to speed up the Perceptron algorithm. Focus of attention speeds up the Perceptron algorithm by lowering the number of features evaluated throughout training and prediction. Whereas the traditional Perceptron evaluates all the features of each example, the Attentive Perceptron evaluates less features for easy to classify examples, thereby achieving significant speedups and small losses in prediction accuracy. Focus of attention allows the Attentive Perceptron to stop the evaluation of features at any interim point and filter the example. This creates an attentive filter which concentrates computation at examples that are hard to classify, and quickly filters examples that are easy to classify.
\end{abstract}
\vspace{-6pt}\section{Introduction}\vspace{-11pt}
Many Online Algorithms base their model update on the margin of each example. Passive online algorithms, such as Rosenblatt's Perceptron \cite{rosenblatt58perceptron} and Crammer et al's online passive-aggressive algorithms \cite{crammer06online}, 
update the algorithm's model only if the value of the margin falls below a defined threshold. These algorithms fully evaluate the margin for each example, even if the model is not to be updated!

The running time of these algorithms is linear either in the number of features, or in the dimensionality of the input space. 
Contemporary models may have thousands of features making running time daunting.
The budgeted learning community addresses this problem by putting a budget on the number of features a classifier can evaluate while learning and while making predictions. Our work stems from the theoretical framework suggested by Ben David and Dichterman \cite{bendavid1998learning}, and is closely related to recent work by Cesa-Bianchi et al. \cite{cesabianchi2010efficient} as well as Reyzin \cite{reyzin10boosting}. 

We differ by the fact that we do not impose a hard budget constraint on the number of features, but rather look at the probability of making decision errors. Decision error are errors that occur when the algorithm stops the feature evaluation process, predicts its outcome, and is wrong. This work extends on previous work by Pelossof et al. \cite{curtailed10pelossof}.

We propose a new method for early stopping the computation of feature evaluations for uninformative examples by connecting the Perceptron algorithm to sequential statistical tests \cite{wald45tests,lan82sequential} (Figure \ref{fig:attentive-perceptron}.) This connection results in a general method that makes margin based learning algorithms attentive, which means that they have the ability to quickly filter uninformative examples.

\vspace{-6pt}\section{The Attentive Perceptron}\vspace{-11pt}
The margin of each example is computed as a weighted sum of feature evaluations. Informative examples are misclassified examples, which force the Perceptron to preform a model update, whereas uninformative examples are correctly classified and therefore ignored by the perceptron.

We break up the feature evaluation for every example in the stream. The breakup of every example allows the Attentive Perceptron to make a decision after the evaluation of each feature about whether the feature evaluation should continue or be stopped.
This decision making process allows us to stop the evaluation of features early on examples with a large partial margin after having evaluated only a few features. For example, examples with a large partial margin are unlikely to have a negative full margin. Therefore, rejecting these examples early achieves large savings in computation.

We define the mathematical setup to derive the stopping conditions for margin evaluation.
Let $X_1,...,X_n$ be weakly dependent random variables. Let a partial sum be defined by $S_i=X_1+...+X_i$ and the remainder sum by $S_{in}=S_n-S_i$. The expectation of a sum is denoted by $ES_i$ and its standard deviation by $std(S_i)$. 

The Perceptron compares the margin (a sum) to a threshold, and updates its model if the margin of the example is negative. We formulate the equivalent sequential decision making process, and drive constant stopping thresholds $\tau$. These thresholds will essentially tell us when it's highly unlikely for the margin to end below the desired importance threshold $\theta$.

The stopping thresholds are derived by requiring that the joint distribution of stopping (and predicting $S_n>\theta$) while the actual full sum satisfies $S_n < \theta$ is less than a required error rate $\delta$
\[P(S_n < \theta,\textit{predict }S_n>\theta) = P(S_n < \theta, S_i > \tau)\le \delta.
\]
We bound the probability of making a decision error 
\begin{eqnarray}
\lefteqn{
P(S_n < \theta , S_i > \tau) \lessapprox P(S_n < \theta, S_i = \tau)} \nonumber\\
&=& P(S_n -ES_n< \theta-ES_n, S_i = \tau) \nonumber\\
&=&P(S_n-ES_n < 2\tau - (\theta-ES_n)) \label{eqn:rw_reflection} \\
&=& P\left(\frac{S_n - ES_n}{std(S_n)} < \frac{2\tau-\theta + ES_n}{std(S_n)}\label{eqn:flat_reflection_standerdized} \right).
\end{eqnarray}
Equation \ref{eqn:rw_reflection} is derived by applying the reflection principle, and equation \ref{eqn:flat_reflection_standerdized} is its standardization.

Since we assume that $X_1,...,X_n$ are weakly independent, the sum $S_n=X_1+...+X_n$ is approximately normally distributed by the Central Limit Theorem. By standardizing $S_n$ we upper bound the probability of making a decision error with the inverse normal cumulative distribution function $\Phi^{-1}$. Therefore, requiring that the probability of making a decision error be less than $\delta$ we get the following equality from equation \ref{eqn:flat_reflection_standerdized}
\begin{equation}
\frac{2\tau-\theta+ES_n}{std(S_{n})} = \Phi^{-1}(1-\delta).
\label{eqn:boundary_inequality}
\end{equation}
The quantities $ES_n$ and $std(S_n)$ can be approximated using the empirical data.

Finally, by solving for the stopping threshold $\tau$ we get from equation \ref{eqn:boundary_inequality}
\begin{equation}
\tau = \frac{1}{2}\left(\theta - ES_n + std(S_n)\Phi^{-1}(1-\delta)\right).
\label{eqn:flat-clt-stst}
\end{equation}
Therefore, examples with partial margin calculations $S_i$ that hit this boundary should be filtered and with probability at least $1-\delta$ determined that their full margin satisfies $S_n>\theta$. 

In summary, we presented a simple test to speed up the Perceptron algorithm by quickly filtering unimportant examples without fully evaluating their features. This results in an algorithm which typically focuses on examples by the decision boundary - the Attentive Perceptron.
\vspace{-7pt}
{
\small
\bibliographystyle{amsplain}
\bibliography{bibdesk}
}
\end{document}